\documentclass[conference]{IEEEtran}
\IEEEoverridecommandlockouts

\usepackage{cite}
\usepackage{amsmath,amssymb,amsfonts}
\usepackage{algorithmic}
\usepackage{graphicx}
\usepackage{textcomp}
\usepackage{xcolor}
\usepackage{booktabs}
\usepackage{multirow}
\usepackage{gensymb}
\usepackage{hyperref}
\usepackage[nameinlink]{cleveref}

\def\BibTeX{{\rm B\kern-.05em{\sc i\kern-.025em b}\kern-.08em\kern-.1667em\lower.7ex\hbox{E}\kern-.125emX}}

\newcommand{\ARCH}{InsTex}

\begin{document}
\title{InsTex: Indoor Scenes Stylized Texture Synthesis}


\author{\IEEEauthorblockN{ Yunfan Zhang$^{1,2}$, Zhiwei Xiong$^{1,2}$, Zhiqi Shen$^{1,2}$, Guosheng Lin$^1$, Hao Wang$^3$, Nicolas Vun$^1$}
\IEEEauthorblockA{\textit{$^1$College of Computing and Data Science, Nanyang Technological University, Singapore} \\
\textit{$^2$Alibaba-NTU Singapore Joint Research Institute, Nanyang Technological University, Singapore} \\
\textit{$^3$The Hong Kong University of Science and Technology (Guangzhou)} \\
\{yunfan001, zhiwei002, zqshen, gslin, aschvun\}@ntu.edu.sg, haowang@hkust-gz.edu.cn}
}

\maketitle

\begin{abstract}
Generating high-quality textures for 3D scenes is crucial for applications in interior design, gaming, and augmented/virtual reality (AR/VR). Although recent advancements in 3D generative models have enhanced content creation, significant challenges remain in achieving broad generalization and maintaining style consistency across multiple viewpoints. Current methods, such as 2D diffusion models adapted for 3D texturing, suffer from lengthy processing times and visual artifacts, while approaches driven by 3D data often fail to generalize effectively. To overcome these challenges, we introduce \ARCH{}, a two-stage architecture designed to generate high-quality, style-consistent textures for 3D indoor scenes. \ARCH{} utilizes depth-to-image diffusion priors in a coarse-to-fine pipeline, first generating multi-view images with a pre-trained 2D diffusion model and subsequently refining the textures for consistency. Our method supports both textual and visual prompts, achieving state-of-the-art results in visual quality and quantitative metrics, and demonstrates its effectiveness across various 3D texturing applications.
\end{abstract}


\section{Introduction}
The generation of high-quality 3D scenes is crucial for visual applications in interior design, gaming, and a wide range of augmented and virtual reality (AR/VR) scenarios. Significant progresses have been achieved through various 3D geometry generative models \cite{mildenhall2020nerf, muller2022instant, stan2023ldm3d, gao2022get3d, lin2022magic3d, EG3D}. However, to fully automate the 3D content creation pipeline, automatic texturing of scenes tailored to diverse application requirements and styles remains a critical area of development.

Recently, 2D diffusion models \cite{luo2021diffusion, ho2020denoisingdiffusionprobabilisticmodels, stablediffusion} have gained more attention due to their excellent performance in generating content. Some approaches, like Latent-Paint \cite{metzer2022latent}, TEXTure \cite{richardson2023texture} and Text2tex \cite{chen2023text2tex}, have extended these 2D models to generate 3D textures by using pre-trained depth-to-image diffusion models guided by text inputs. However, these methods have certain limitations. MVDiffusion \cite{tang2023mvdiffusion} trains a multi-view diffusion model with cross-attention across multiple branches to ensure view consistency in indoor scenes. Whereas, MVdiffusion can only generate panoramic views for indoor scenes, and the lower quality of the textured meshes limits its applicability in downstream tasks. SceneTex \cite{chen2023scenetexhighqualitytexturesynthesis}, for example, uses a multi-resolution texture field and optimizes the texture through Variational Style Diffusion (VSD) \cite{wang2023prolificdreamer} with multiple RGB renderings. This fine-tuning process can take over a day to finish and often produces artifacts in the background, as shown in Figure \ref{fig:baselines}. On the other hand, methods that are trained directly on 3D data, like PointUV \cite{yu2023texturegeneration3dmeshes} and Mesh2tex \cite{bokhovkin2023mesh2texgeneratingmeshtextures}, create textures by considering the full geometry of specific 3D objects. However, these methods often lack of generalization, making it difficult to apply them to a wide range of 3D objects beyond their training datasets or to produce varied textures based on different text or visual prompts.

\begin{figure*}[t]
\centering
\includegraphics[width=\textwidth]{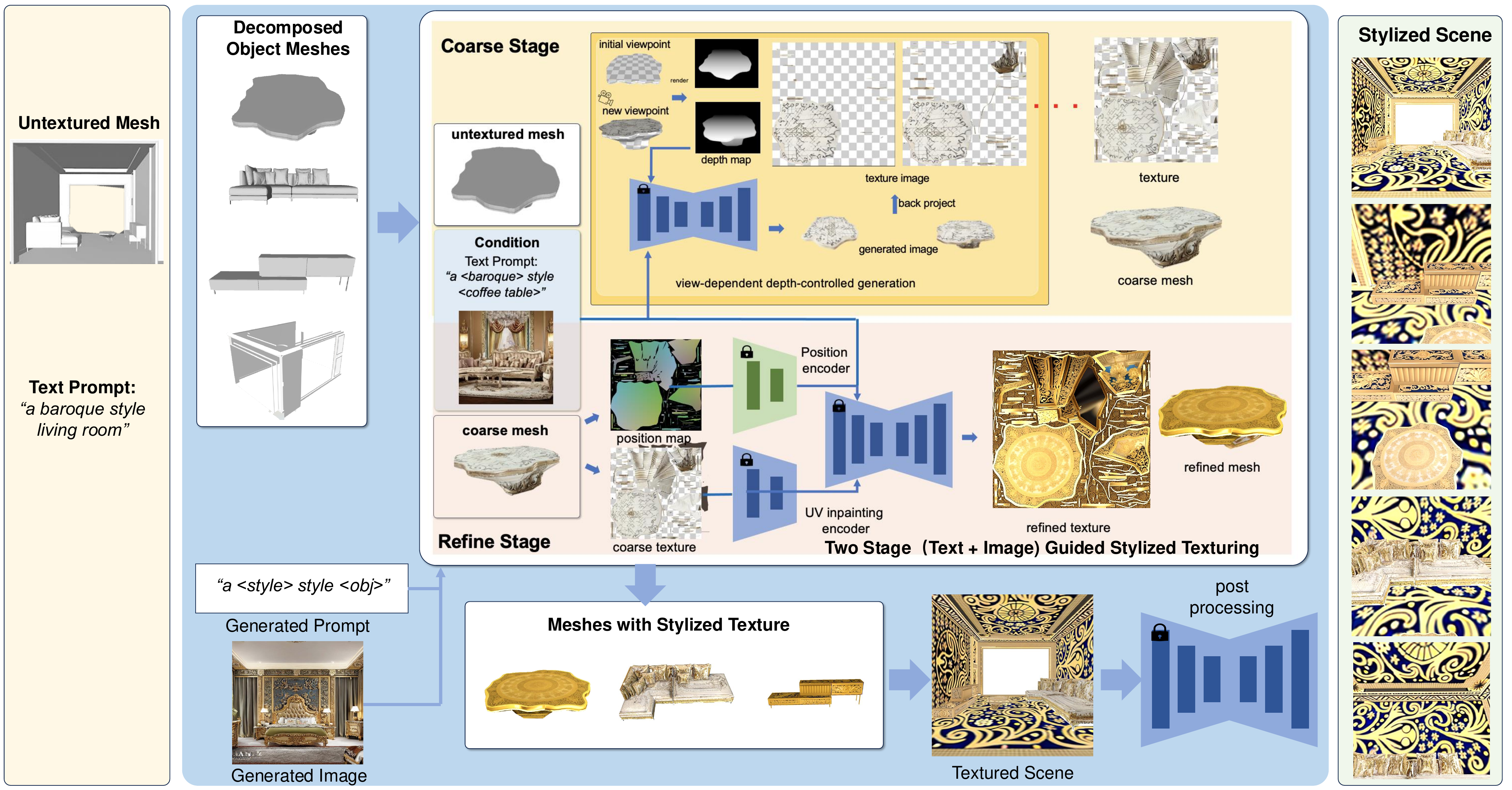}
\caption{InsTex Pipeline: The proposed InsTex pipeline starts with an untextured scene mesh and a style prompt, such as "baroque bedroom" in this case. It then generates a high-quality, visually pleasing, and stylized "baroque" living room, with textures that are conditioned on the given prompt.}
\label{fig:pipeline}
\end{figure*}

There are two main challenges in 3D indoor scene generation. The first is achieving broad generalization across different objects using varied prompts or image guidance. The second is ensuring consistency in style and appearance from multiple viewpoints, which is difficult because 2D models do not fully understand the shapes of 3D objects.

To address the challenges mentioned earlier, we propose \ARCH{}, a novel architecture for generating high-quality, style-consistent textures for indoor scene meshes using depth-to-image diffusion priors, as illustrated in Figure \ref{fig:pipeline}. To achieve rich, high-quality textures from diverse prompts while maintaining style consistency across different views, our method decomposes the scene into individual objects and textures them based on text prompts and a generated global image guidance. During the texturing stage, we first employ a progressive texture generation approach in the coarse phase by sampling multi-view images from a pre-trained, depth-aware 2D image diffusion model and back-projecting these images onto the 3D mesh surface to form an initial texture map. This is followed by a refinement phase, which enhances the textures by filling in any missing areas and fine-tuning the overall texture quality from a holistic perspective. Finally, we recompose the scene by post-processing the entire scene. We demonstrate that the proposed pipeline can effectively generate diverse, high-quality, and visually appealing textures. Additionally, we emphasize the strengths of \ARCH{}, which leverages the powerful image generation and prompt guidance capabilities of pre-trained image generative models for 3D object texturing.

To summarize, our contributions are threefold:
\begin{itemize}
    \item We introduce a two-stage pipeline for indoor scene stylization, where each object is individually textured using a coarse-to-fine approach to ensure detailed refinement.
    \item We ensure scene consistency through global style guidance, managed by a comprehensive global scene image.
    \item Our method supports both textual and visual prompts as inputs and achieves state-of-the-art performance in texturing 3D indoor scenes, surpassing recent works in both visual quality and quantitative analysis. 
\end{itemize}

\section{The proposed approach}

Our goal is to generate stylized textures for a 3D indoor scene based on a given prompt. Firstly, the indoor scene is decomposed into individual objects. These objects are projected into canonical space by translating the center to the origin and the dimension into the unit cubic bounding box for object-level texturization. Secondly, the coarse-to-fine object-level texturing pipeline is adopted. During the texturing stage, to achieve style consistency across different objects, we adopt both a stylized prompt and a generated \textit{global} stylized image as the dual conditions. The \textit{global} image is generated using a text2image \cite{stablediffusion} generation pipeline. To synthesize the appearance of the input geometry, we project the generated 2D views onto the texture space of a normalized 3D object with proper UV parameterization. Thirdly, the indoor scene is re-composed based on the position information captured during a de-composition stage. Lastly, a post-processing stage for the whole indoor scene is conducted to further enhance the integrity and consistency.


\subsection{Scene Object Decomposition}
The indoor scene mesh can be subdivided into individual objects using various methods such as 3D segmentation pipelines \cite{Roy_2023, lei2023meshconvolutioncontinuousfilters}, or by projecting into 2D space and use 2D segmentation pipelines \cite{kirillov2023segment} as well. The main objective of this stage is to obtained the 3D location coordinates and bounding box information. In our pipeline, we use the pre-processed scene mesh following \cite{chen2023scenetexhighqualitytexturesynthesis}.


\subsection{Coarse Texture Generation}
We adopt the diffusion model \cite{stablediffusion, ho2020denoisingdiffusionprobabilisticmodels} as the \textit{prior} knowledge in order to achieve stylization in a diverse manner. In our pipeline, the depth2image model is used to generate a high-quality image of the object based on the depth map from a specific single view.

The objects are firstly translated into their canonical position. Predefined viewpoints are defined to cover every part of the objects, thus we choose to view at the object center which is also the world coordination origin from a distance of 1 with the elevation angle 15\degree{} with respect to xz-plane. The azimuth angle is rotated from 0\degree{} to 360\degree{} in step size of 45\degree{}. Lastly, the top and bottom view are also included to avoid any missing parts. Whereas, a typical room mesh contains two sides. We are only interested in the inner surface, thus, the viewing points chosen are slightly different from those we choose for objects. The viewpoints are chosen to look from the object center and rotate the viewpoint around the up-axis from 0\degree{} to 360\degree{}. The top and bottom-down views are also included.

The texture generation starts from an initial viewpoint v0. We render the object to a depth map D0 and a generation mask M0, and then use the depth to image diffusion model with D0 as input and M0 as extra guidance to generate a colored image, which is back-projected to the visible part of the texture T0. In the subsequent steps, we progressively diffuse the colored images and back-project them to the texture through the predefined viewpoints. 

It is obvious that by directly inpainting the missing regions on a mesh surface often results in inconsistency issues. The issue is mainly caused by the stretched artifacts that occur when the 2D views are projected back onto the curved surface of the mesh. Therefore, we design a dynamic view partitioning strategy to guide the inpainting process with respective generation objectives M and denoising strengths $\gamma$. This strategy divides the viewpoints covering the entire object into different regions using a generation mask, which directs the depth-aware inpainting model.

Our pipeline begins with the front view of the object, from which a depth map is obtained by rendering the mesh. This depth map is then used to condition a depth-aware image inpainting pipeline, generating a front-view image that includes mesh geometry information. Additionally, we incorporate a text \textit{prompt} and a previously generated \textit{global} image to ensure the texture is accurate and of high quality. The generated image is subsequently back-propagated into UV-map space to update the UV-map.

After updating the UV-map, a new viewpoint is selected to render a new image of the mesh. This new section is segmented into three regions: "generate," "update," and "keep." The "generate" region consists of new areas visible from the current viewpoint, where new appearances are created by denoising white Gaussian noise. This technique refines the previously generated image content by synthesizing textures that fill in the missing regions, ensuring the texture closely matches the mesh geometry and viewpoint. The "update" regions, which lie mainly at the boundaries between the previous and new viewpoints, involve enhancing existing textures by denoising partially noised segments. The "keep" region retains its texture from the current view.

Dynamic masks are generated for each viewpoint to guide the model's focus. By iterating through predefined viewpoints, the coarse texture can be generated while maintaining geometric and stylistic consistency.

\subsection{Texture Refinement}
The coarse texture still has some artifacts due to the object self-occlusion and different diffusion update strengths in different regions. To get a better quality texture, we propose a refine stage to refine the texture in UV space. Whereas, the UV map is in a semi-continuous state, meaning the element in a part is continuous and becomes discontinuous when reaching the part boundary. In this case, those diffusion models well pre-trained mainly in normal images which are all continuous in nature, cannot be used directly. To this end, a diffusion process guided by adjacency information of texture fragments is proposed. Here, the 3D adjacency information of texture fragments is represented as the position map in UV space $R\in H\times W\times 3$, where each element is a 3D point coordinate. In this case, this position map is a analogue to the positional encoder in transformer design, serving as a bridge to link the part boundary in the UV map continuously. The position map can be easily obtained through UV mapping of the 3D point coordinates. To fuse the 3D adjacency information during the diffusion process, we add an individual position map encoder to the pre-trained image diffusion model. Following the design principle of ControlNet, the new encoder has the same architecture as the encoder in the image diffusion model and is connected to it through a zero-convolution layer.

We can simultaneously use the position encoder and UV map encoders to perform the refinement tasks in UV space. The UV inpainting is used to refine any artifact within the UV plane, which can avoid self-occlusion problems during rendering. Moreover, to achieve the style-consistency in all objects, we also adopt the IP-adaptor which takes in the global style image generated in the coarse stage. In our refinement stage,  the coarse texture map can be refined to achieve high-resolution and diverse UV texture maps.

\begin{figure*}[t]
    \centering
    \includegraphics[width=\textwidth]{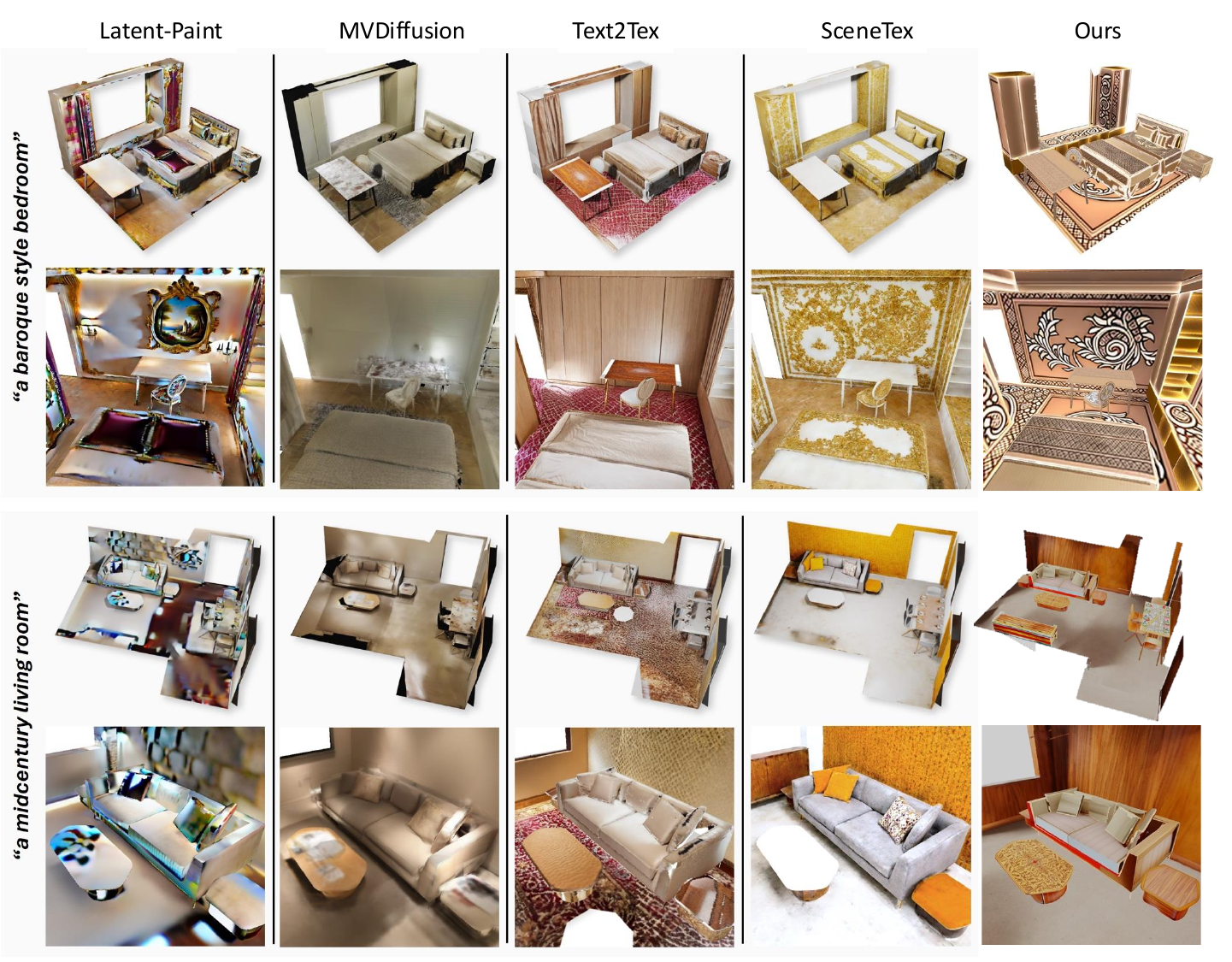}
    \caption{ 
    \textbf{Qualitative comparisons. }
    Latent-Paint~\cite{metzer2022latent} suffers from issues of over-saturation and the hallucination of scene elements, while MVDiffusion~\cite{tang2023mvdiffusion} produces blurry textures that do not accurately reflect the input prompts. Text2Tex~\cite{chen2023text2tex} encounters difficulties in maintaining style consistency across all instances. SceneTex \cite{chen2023scenetexhighqualitytexturesynthesis} shows some artifacts in both floor and wall. Our method generates high-quality textures while preserving overall style consistency throughout the scenes.
    }
    \label{fig:baselines}
\end{figure*}

\subsection{Scene Re-composition and Post-processing}
With all textured rooms and objects, the scene is re-composed using the previously recorded information such as scales and positions during the translation. With all objects textured and in place, we pass the scene texture map through our final process stage which is also a diffusion model to further remove any artifact to archive the overall consistency in our final mesh.

\section{EXPERIMENTAL EVALUATION}
\subsection{Implementation Details}
We mainly rely on Stable Diffusion v1.5~\cite{stablediffusion} as our backbone. Our full implementation uses the PyTorch framework, with PyTorch3D~\cite{ravi2020pytorch3d} for rendering and texture projection.

For global reference image generation, we use the pre-trained stable-diffusion-v1-5 diffusion model from runwayml with all default pre-settings. The standard text prompt is in the form of \textit{a \textless{}style\textgreater{} style \textless living room/bedroom\textgreater }. The depth-controlled generation model is a pretrained ControlNet model based on stable-diffusion-v1-5 as well. In the proposed object-specific texturization method, the text prompt is dynamically adjusted to replace the general room style descriptor with the name of the target object. For example, the prompt ``a Baroque style coffee table” is constructed to correspond directly to the object being textured, ensuring semantic and stylistic alignment. The global style image is processed using an IP-Adapter to extract relevant features, while the text prompt is encoded into an embedding space via the CLIP text encoder. These embeddings are subsequently concatenated with depth image embeddings to form a unified conditioning input for the texturization pipeline.

In the coarse stage, a diffusion-based model is employed to generate and enhance textures. For regions designated as ``generate," 50 diffusion steps are utilized to create high-quality images that align closely with the input text prompt. For ``update" regions, the diffusion steps are reduced to 10, focusing on denoising and refining specific portions to enhance local details while maintaining the coherence of the overall texture. In the refinement stage, a ControlNet-based inpainting model is applied, conditioned on UV positional maps. These maps encode continuous positional information within the range of 0 to 1, providing precise spatial context for texture application. During this stage, an intermediate mask is generated to differentiate foreground and background elements, ensuring accurate and context-aware inpainting of the texture.

\subsection{Dataset}

We conduct experiments on a subset of textured meshes from the 3D-FRONT~\cite{fu20203dfront} dataset following the experiment settings of SceneText~\cite{chen2023scenetexhighqualitytexturesynthesis}. There are 10 3D-FRONT scenes with 2 different text prompts for each scene.

\subsection{Evaluation Metrics}
CLIP score~\cite{mohammad2022clip} is a widely adopted metric in research and applications involving multi-modal AI systems. It is a metric used to evaluate how well a generated image aligns with a given textual description, by computing the embeddings for both images and text, enabling a direct comparison of their alignment in a shared semantic space. In this experiment, we render the indoor scene from 20 preset view points into 20 2D images and calculate the average CLIP score of these images with the text prompt as our final results.

The Inception Score (IS)~\cite{smith2017improved} is a widely used metric for evaluating the quality of images generated by models. It assesses two key aspects of the generated images: the diversity of the generated content and the quality of the images, which is determined by how well the model captures meaningful, distinct classes. 

Following the experiment settings of SceneTex~\cite{chen2023scenetexhighqualitytexturesynthesis}, we calculate CLIP score~\cite{mohammad2022clip} and IS~\cite{smith2017improved} to measure the fidelity with input prompts and texture quality, respectively. We additionally report the User Study results from 100 participants about the Visual Quality and Prompt Fidelity on a scale of 1-5, in which 1 is the worst and 5 is the best quality.

\subsection{Qualitative Results}
\begin{figure}[t]
    \centering
    \includegraphics[width=\columnwidth]{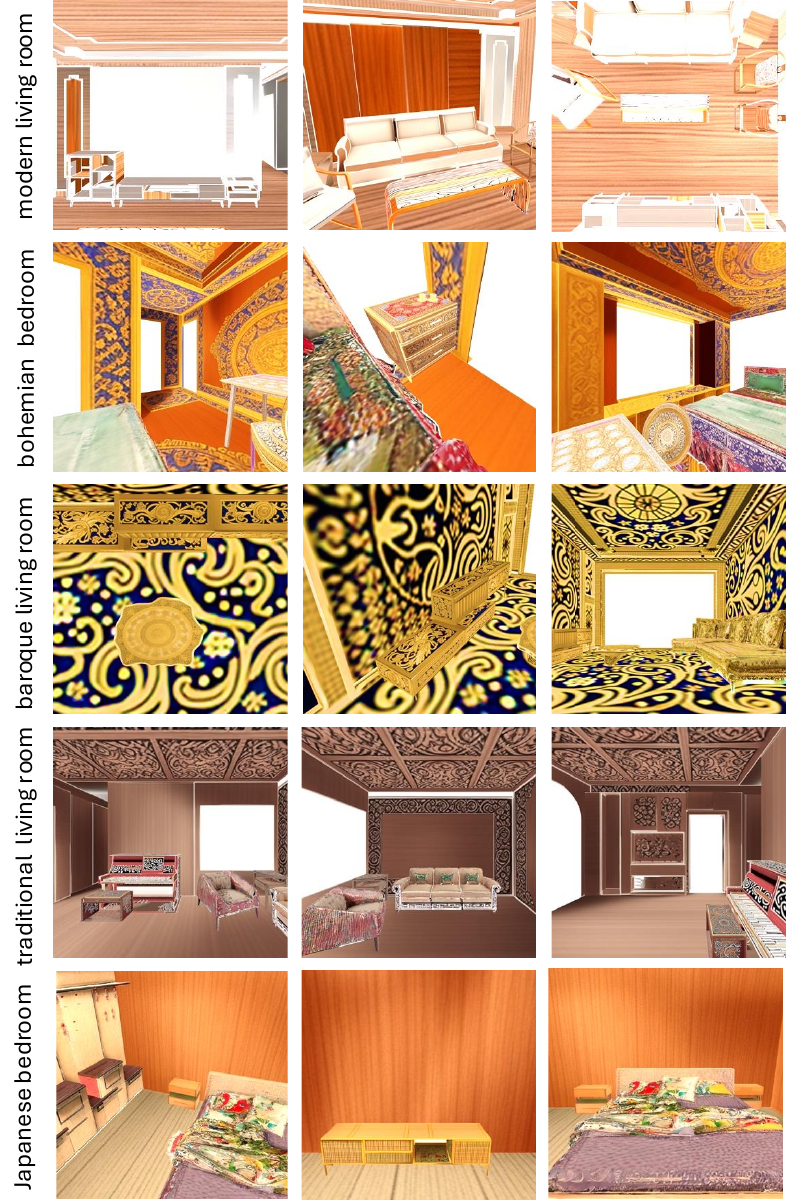}
    \caption{Synthesized textures for 3D-FRONT scenes. Our method generates high-quality style-coherent textures, and reflects the iconic traits in the prompts.
    }
    \label{fig:results}
\end{figure}

We present the qualitative comparisons in Figure \ref{fig:baselines}. Latent-Paint struggles with over-saturation and introduces non-existent elements, such as a large frame on the wall. These unrealistic textures result from inaccurate geometric cues and a mismatch between the optimized latent representation and the final texture. MVDiffusion~\cite{tang2023mvdiffusion} generates textures that are generally smooth but appear blurry and dim. It also fails to capture the distinctive characteristics described in the prompts, such as "baroque" and "luxury." Text2Tex~\cite{chen2023text2tex} produces realistic textures for individual objects but lacks global style consistency across the scene. SceneTex \cite{chen2023scenetexhighqualitytexturesynthesis} delivers sub-optimal results, especially on the wall and bed sheet, creating a somewhat chaotic appearance. The wall, in particular, features an unrealistic "water-wave" pattern, which is also evident in other cases. In contrast, our method generates high-quality textures with consistent styles both within and across objects, accurately capturing the representative traits in the prompts with high fidelity. Additionally, we visualize the results of our texture synthesis for various 3D-FRONT~\cite{fu20213dfront} scenes and input prompts in Figure \ref{fig:results}, showcasing the superior texture quality and fidelity achieved by our approach.

\subsection{Quantitative Results}
\begin{table}[t]
    \caption{We report the 2D metrics and User Study results for quantitative comparisons, including: CLIP score (CLIP)~\cite{radford2021learning}, Inception Score (IS)~\cite{smith2017improved}, Visual Quality (Visual Quality), and Prompt Fidelity (PF). We show that our method produces textures with the highest quality.
    }
    \centering
    \begin{tabular}{l cc cc}
        \toprule
            \multirow{2}{*}{Method} & \multicolumn{2}{c}{2D Metrics} & \multicolumn{2}{c}{User Study} \\
                            \cmidrule(l{2pt}r{2pt}){2-3} \cmidrule(l{2pt}r{2pt}){4-5}
        & CLIP $\uparrow$ & IS $\uparrow$ & VQ $\uparrow$ & PF $\uparrow$\\
        \midrule
        Latent-Paint~\cite{metzer2022latent} & 18.37 & 1.96 & 1.57 & 2.11 \\
        MVDiffusion~\cite{tang2023mvdiffusion} & 18.47 & 2.83 & 3.09 & 3.12 \\
        Text2Tex~\cite{chen2023text2tex} & 20.83 & 2.87 & 2.62 & 3.04 \\
        SceneTex ~\cite{chen2023scenetexhighqualitytexturesynthesis} & 22.18 & 3.33 & 4.40 & 4.29 \\
        \midrule
        \ARCH (Ours) & \textbf{24.18} & \textbf{3.82} & \textbf{4.70} & \textbf{4.82} \\
        \bottomrule
    \end{tabular}
    \label{tab:quantitatives}
\end{table}
In Table. \ref{tab:quantitatives}, we present the quantitative comparisons with the previous SOTA methods in indoor scene texture synthesis. Following \cite{chen2023scenetexhighqualitytexturesynthesis}, we report the CLIP Score and Inception Score (IS) to evaluate the quality and diversity of the generated textures. Our method significantly outperforms all baselines, highlighting its superior ability to produce high-quality textures for a wide range of objects across various categories.

Table. \ref{tab:timing} is a comparative table illustrating the processing time required for scene generation conditioned on the given prompt. Both Latent-Paint and MVDiffusion primarily generate panoramic images within their core models, followed by post-processing using the tool TSDF Fusion to produce the textured mesh. The reported timing represents the total duration, encompassing both inference and post-processing stages. Our pipeline achieves the \textbf{fastest} processing time.
\begin{table}[t]
    \caption{The Processing time for a typical living room containing a bed, a side table, a wardrobe, and a dressing table set. The GPU used in the experiments is Nividia V100, 32G.
    }
    \centering
    \begin{tabular}{l c}
        \toprule
        Method & Timing (hours) \\
        \midrule 
        Latent-Paint & 10 \\
        MVDiffusion& 10 \\
        Text2Tex& 6 \\
        SceneTex& 48 \\
        \midrule
        \ARCH (Ours) & \textbf{2}  \\
        \bottomrule
    \end{tabular}
    \label{tab:timing}
\end{table}

User Study: We conduct a user study to evaluate the overall quality of the generated textures and their adherence to the input text prompts. For this study, we randomly select five meshes and their corresponding text prompts. These meshes are textured using both our method \ARCH{} and baseline models and presented to users in a random order. Each object is shown with full-texture details through a 360\degree{} rotation. Participants assess the textures based on two criteria: (1) overall quality and (2) fidelity to the text prompt, using a 1 to 5 scale. We collected evaluations from 100 users, as summarized in Table. \ref{tab:quantitatives}. It shows the average scores for each method across all prompts. Our approach significantly outperforms the baseline methods in both overall quality and fidelity to the text prompt.

\subsection{Ablation Study}
We conduct the some ablation studies to demonstrate the effectiveness of our individual components.

\begin{figure}[t]
    \centering
    \includegraphics[width=\columnwidth]{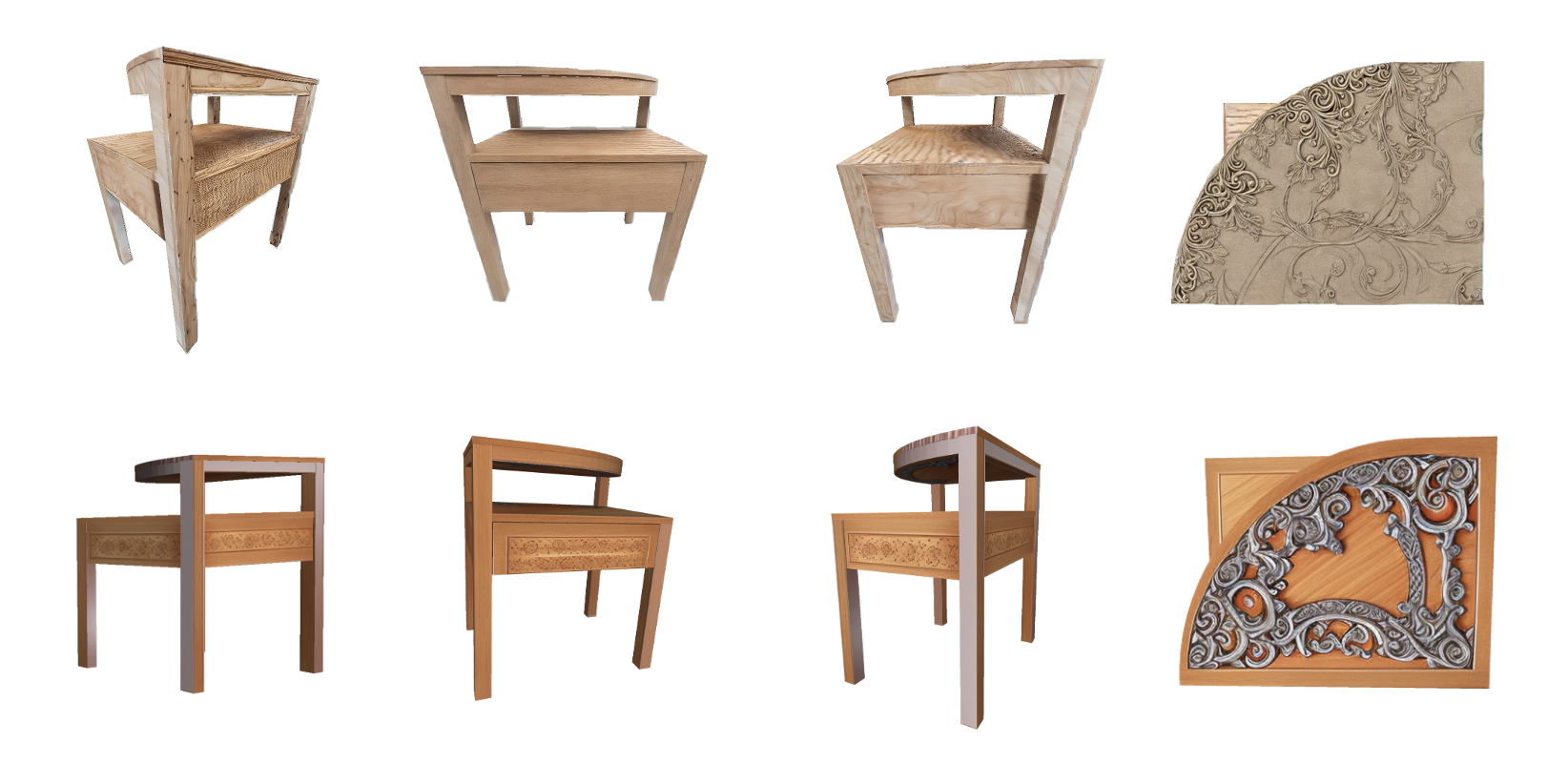}
    \caption{The comparison of the textured table. The first raw shows the coarse stage outputs, the second row shows the the refinement stage outputs. After refinement, the texture quality is improved with better visual quality and the missing parts like the missing joints are in-painted as well.}
    \label{fig:instex_coarse_refine}
\end{figure}

Figure \ref{fig:instex_coarse_refine} illustrates the textured table at both the coarse and refinement stages. The first row of Figure \ref{fig:instex_coarse_refine} presents the outputs from the coarse stage, while the second row displays the results from the refinement stage. After refinement, the texture quality is significantly improved, with enhanced visual fidelity. Additionally, missing parts, such as the absent joints, are successfully in-painted, further enhancing the overall coherence of the texture.

\begin{figure}[t]
    \centering
    \includegraphics[width=\columnwidth]{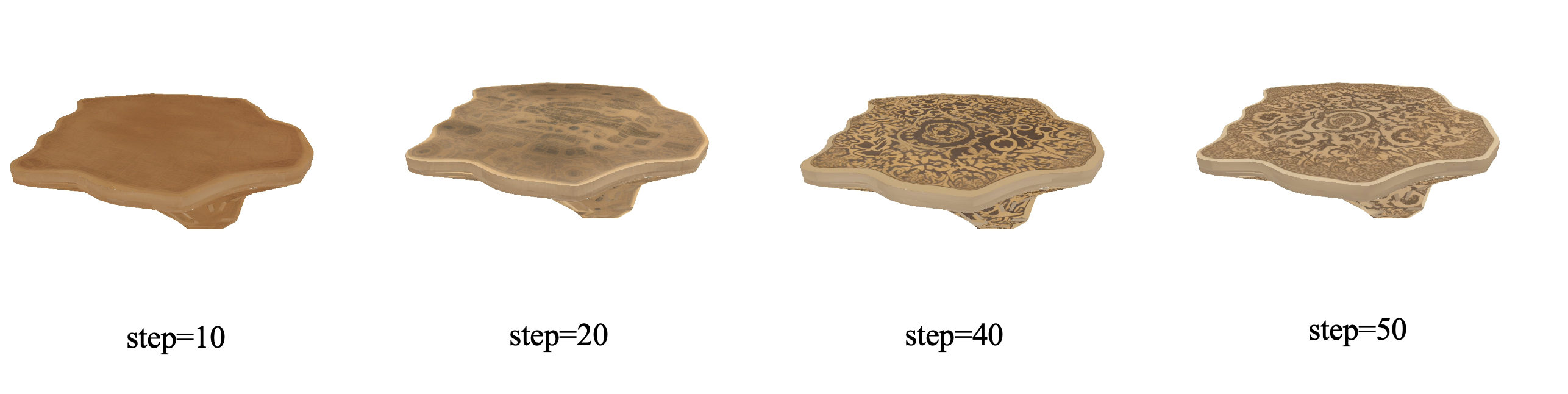}
    \caption{The texturization process for the \textit{Baroque} coffee table was evaluated under varying diffusion step settings during the coarse stage. When utilizing 50 diffusion steps, the generated textures exhibited vivid details and achieved optimal alignment with the distinctive Baroque style, demonstrating the effectiveness of this configuration in capturing intricate stylistic elements.}
    \label{fig:instex_diffusion_steps}
\end{figure}

We also conduct experiments to evaluate the impact of diffusion steps on the quality of the textured output. As shown in Figure \ref{fig:instex_diffusion_steps}, a \textit{Baroque} coffee table is textured. In this experiment, the lesser the steps, the more details cannot be denoised. When utilizing 50 diffusion steps, the generated textures exhibited vivid details and achieved optimal alignment with the distinctive Baroque style, demonstrating the effectiveness of this configuration in capturing intricate stylistic elements.

\section{Conclusion}
We introduce \ARCH{}, a novel method for effectively generating high-quality and style-consistent textures for indoor scenes using depth-to-image diffusion priors. At its core, the global style awareness for each instance results in a style-coherent appearance in the target scene. Extensive analysis shows that \ARCH{} enables various and accurate texture synthesis for 3D-FRONT scenes.

\bibliographystyle{IEEEtran}
\bibliography{main}

\end{document}